\documentclass[11pt]{article}

\usepackage[final]{acl}

\usepackage{times}
\usepackage{latexsym}
\usepackage{booktabs}
\usepackage{tabularx}
\usepackage{float}
\usepackage{amssymb} 

\usepackage{tikz}
\usetikzlibrary{positioning, fit, backgrounds, arrows.meta, calc, shapes.geometric}
\usepackage{booktabs, multirow, colortbl, pifont, xcolor, makecell}
\usepackage{hyperref} 

\hypersetup{
    colorlinks=true,
    linkcolor=blue,
    urlcolor=blue,
}
\usepackage{amsmath}
\usepackage{subcaption}
\usetikzlibrary{positioning, arrows.meta, fit, calc, backgrounds}

\definecolor{corecol}{RGB}{41,98,178}
\definecolor{corelite}{RGB}{210,225,255}
\definecolor{optcol}{RGB}{90,90,90}

\definecolor{corecol}{RGB}{41,98,178}
\definecolor{arcol}{RGB}{200,80,20}
\definecolor{disccol}{RGB}{120,40,170}
\definecolor{elfcol}{RGB}{15,130,90}
\definecolor{optcol}{RGB}{90,90,90}

\usepackage[T1]{fontenc}

\usepackage[utf8]{inputenc}
\usepackage{pgfplots}
\usepgfplotslibrary{groupplots}
\pgfplotsset{compat=1.18}
\usepackage{microtype}
\usepackage{listings,xcolor}

\definecolor{lbg}{RGB}{245,247,252}
\definecolor{lkw}{RGB}{25,75,155}
\definecolor{lst}{RGB}{135,55,0}
\definecolor{lcm}{RGB}{90,90,90}
\definecolor{lnu}{RGB}{100,0,140}
\definecolor{lfr}{RGB}{190,200,220}

\lstdefinestyle{dxpy}{
  language        = Python,
  basicstyle      = \ttfamily\fontsize{7.0}{8.5}\selectfont,
  backgroundcolor = \color{lbg},
  commentstyle    = \color{lcm}\itshape,
  keywordstyle    = \color{lkw}\bfseries,
  stringstyle     = \color{lst},
  morekeywords    = {True,False,None},
  showstringspaces= false,
  breaklines      = true,
  breakatwhitespace=true,
  columns         = flexible,
  keepspaces      = true,
  frame           = single,
  framesep        = 2.5pt,
  rulecolor       = \color{lfr},
  xleftmargin     = 4pt,
  xrightmargin    = 4pt,
  aboveskip       = 3pt,
  belowskip       = 2pt,
}
\usepackage{inconsolata}

\usepackage{graphicx}

\title{\textsc{DantinoX}: A Unified  Framework \\ for Multi-Paradigm Language Modeling}

\author{
  Marco Simoni \quad Aleksandar Fontana \quad Giulio Rossolini \quad Andrea Saracino \\
  Department of Excellence in Robotics and AI, TeCIP \\
  Scuola Superiore Sant'Anna, Pisa 
}

\begin{document}

\definecolor{dxGreen}{RGB}{27,94,32}
\definecolor{dxOrange}{RGB}{230,81,0}
\definecolor{dxRed}{RGB}{183,28,28}
\definecolor{dxGray}{RGB}{176,176,176}
\definecolor{dxRowHL}{RGB}{232,245,233}
\definecolor{dxRowAlt}{RGB}{249,249,252}

\maketitle
\begin{abstract}
Language generation research increasingly spans three paradigms: autoregressive decoding, discrete masked diffusion, and continuous flow-matching. Comparing them is difficult because each lives in a separate codebase, so measured differences often reflect implementation details rather than the paradigms themselves. We present \textsc{DantinoX}, an open-source JAX/Flax library in which a single modular Transformer backbone serves all three paradigms. 
Switching the generation paradigm, attention mechanism, or hardware topology requires only a configuration change, while the backbone architecture, tokenizer, initialization strategy, and training infrastructure remain consistent. This enables controlled cross-paradigm comparisons within one API for training, streaming inference, and benchmarking.
Using its own pipeline, we evaluate generation quality and attention combinations and profile inference latency, throughput, and energy through a hardware roofline analysis. \textsc{DantinoX} is released under the MIT license and installable via \texttt{pip}, with code, documentation, notebooks, and a demonstration video publicly available\footnote{\href{https://pypi.org/project/dantinox/}{Installation Package} ~|~  \href{https://youtu.be/1u5-AieDzIc}{Demo video}}.
\end{abstract}

\section{Introduction}
Language modeling is no longer synonymous with left-to-right decoding. Alongside the dominant Autoregressive (AR) paradigm~\cite{gpt3}, Masked Diffusion~\cite{austin2021structured, diffusion_model_1} and continuous flow-matching~\cite{hu2026elf} are viable alternatives~\cite{survey_llm}, differing in training objective, attention masking, decoding procedure, and inference cost. Yet comparing them empirically is challenging~\cite{dodge2019show}. Each paradigm has grown its own software ecosystem, while production frameworks remain optimized for AR inference alone~\cite{shoeybi2019megatron}. A researcher comparing diffusion against an AR baseline must port models across incompatible repositories, and any difference they measure may reflect tokenizers, initialization, or training-loop details rather than the paradigm itself. What is missing is a shared substrate on which paradigms can be trained, compared, and served under identical conditions~\cite{dodge2019show}.

We present \textsc{DantinoX}, an open-source library providing this substrate. Built in JAX/Flax~\cite{jax2018github, flax2020github}, it fully leverages JAX's functional programming model and composable transformations. Its modular Transformer backbone supports all three paradigms via simple configuration, offering a unified interface for the entire training, evaluation, and serving lifecycle. The backbone is highly composable in its attention mechanisms, feed-forward layers, LoRA adapters~\cite{hu2022lora}, and data/tensor parallelism~\cite{adler2024nemotron}, and includes an integrated benchmarking suite.

\textsc{DantinoX} is intended to serve a broad audience: researchers, by enabling ablations through configuration changes rather than model-code edits; educators, by providing an accessible and unified codebase for teaching multi-paradigm NLP; and practitioners in small enterprises, by offering reusable pipelines that reduce the need to build such infrastructure from scratch~\cite{ai-for-sme, ai-for-sme-1, ai-for-sme-2}.
The library is MIT-licensed, with code, documentation and configurations available on GitHub~\footnote{\href{https://github.com/winstonsmith1897/DantinoX}{Github}}. We validate it with three evaluation suites (Section~4):
a cross-check of the paradigm implementations against other
 codebases~\cite{zhou2026dllm, xlm2026eacl}, generation quality (MAUVE~\cite{pillutla2021mauve},
perplexity, diversity, conditional BLEU~\cite{papineni-etal-2002-bleu})
across all paradigm-attention combinations, and inference efficiency.

\begin{table*}[t]
\centering
\footnotesize
\setlength{\tabcolsep}{4pt} 
\renewcommand{\arraystretch}{1.15} 
\newcommand{\cmark}{{\color{dxGreen}\ding{51}}}
\newcommand{\xmark}{{\color{dxGray}\ding{55}}}
\begin{tabular}{@{} l l @{\hspace{6pt}} c c c @{\hspace{6pt}} l @{\hspace{6pt}} c c c @{}}
\toprule
\multirow{2}{*}{\textbf{Framework}} &
\multirow{2}{*}{\textbf{Ecosystem}} &
\multicolumn{3}{c}{\textbf{Paradigm Support}} &
\multirow{2}{*}{\textbf{Attention Variants}} & 
\multicolumn{3}{c}{\textbf{LLM Infrastructure}} \\
\cmidrule(lr){3-5}\cmidrule(lr){7-9}
 & & \textbf{AR} & \textbf{Discrete} & \textbf{Contin.}
 & & \textbf{LoRA} & \textbf{Multi-GPU} & \textbf{Bench. Suite} \\ 
\arrayrulecolor{black!25}\midrule\arrayrulecolor{black}
\rowcolor{dxRowAlt}
HuggingFace   & PyTorch~/~JAX       & \cmark & \xmark & \xmark & MHA, GQA, MLA          & \cmark & \cmark & \xmark \\
MaxText       & JAX~/~Flax          & \cmark & \xmark & \xmark & MHA, GQA               & \cmark & \cmark & \xmark \\
\rowcolor{dxRowAlt}
Levanter      & JAX~/~Flax          & \cmark & \xmark & \xmark & MHA, GQA               & \cmark & \cmark & \xmark \\
OpenLM        & PyTorch             & \cmark & \xmark & \xmark & MHA                    & \xmark & \cmark & \xmark \\
\rowcolor{dxRowAlt}
torchtune     & PyTorch             & \cmark & \xmark & \xmark & MHA, GQA               & \cmark & \cmark & \xmark \\
Fairseq       & PyTorch             & \cmark & \xmark & \xmark & MHA                    & \xmark & \cmark & \xmark \\
\rowcolor{dxRowAlt}
xLM           & PyTorch             & \cmark & \cmark & \xmark & MHA                    & \xmark & \cmark & \xmark \\
dLLM          & PyTorch             & \xmark & \cmark & \xmark & MHA                    & \cmark & \cmark & \xmark \\
\midrule
\rowcolor{dxRowHL}
\textbf{\textsc{DantinoX} (Ours)}  & \textbf{JAX~/~Flax} & \cmark & \cmark & \cmark & \textbf{MHA, GQA, MLA} & \cmark & \cmark & \cmark \\
\bottomrule
\end{tabular}
\caption{\small{Comparison of \textsc{DantinoX} with existing NLP generative modeling frameworks.}}
\label{tab:framework_comparison}
\vspace{-3mm} 
\end{table*}

\section{Related Work}

Language modeling frameworks split sharply along the AR/non-AR boundary, trading scale for flexibility. Production-oriented libraries such as Hugging Face \texttt{transformers}~\cite{wolf2020huggingfaces}, Megatron-LM~\cite{shoeybi2019megatron}, Nanotron~\cite{nanotron2023}, and MaxText~\cite{maxtext2023} provide highly optimized infrastructure for large-scale training, but their generation and decoding pipelines assume left-to-right causal inference.
A second wave of research-centric frameworks favors modularity and reproducibility over raw scale: Levanter~\cite{levanter2023} and OpenLM~\cite{openlm2023} target reproducible AR pretraining, while torchtune~\cite{torchtune2024} focuses on AR post-training and fine-tuning. All three remain confined to the AR paradigm, requiring invasive changes to their core APIs and attention mechanisms to support non-autoregressive objectives such as masked diffusion or continuous flow-matching. Fairseq~\cite{ott2019fairseq} offered early support for parallel decoding via non-autoregressive translation, but never extended this to continuous-space objectives or modern attention variants.
Non-autoregressive libraries close this gap only partially. \texttt{xLM}~\cite{xlm2026eacl} supports an AR baseline alongside discrete masked diffusion on a shared backbone, but not continuous flow-matching. dLLM~\cite{zhou2026dllm} targets discrete diffusion variants (masked, block, edit-based) for models such as LLaDA~\cite{nie2024llada} and Dream~\cite{ye2025dream7bdiffusionlarge}, without a matched AR baseline on the same backbone. In short, no existing framework unifies AR, discrete-diffusion, and continuous flow-matching training on a single backbone, forcing cross-paradigm studies to reconcile heterogeneous implementations that obscure genuine algorithmic variance.

\textsc{DantinoX} aims to close this gap. As Table~1 summarizes, it combines all three paradigms on a single JAX/Flax backbone, alongside the attention variants (MHA, GQA, MLA), LoRA fine-tuning, multi-GPU scaling, and benchmarking infrastructure that existing tools tend to offer only in isolation. Switching paradigm, attention mechanism, or hardware topology requires only a configuration change rather than a new codebase, letting researchers and practitioners work within the same API.

\section{\textsc{DantinoX} Architecture}

Evaluating text generation paradigms such as autoregressive, discrete diffusion, and flow-matching requires shared architectures and training setups. When methods use different codebases, implementation details can hide algorithm behaviour. \textsc{DantinoX} addresses this by separating the model backbone from the generation method. The core model, or \textit{backbone}, is built from interchangeable components, such as attention, feed-forward layers, and normalisation. Users can then select the generation paradigm with a single configuration setting. 

\subsection{Backbone}
To ensure a controlled comparison across paradigms, the shared backbone (Figure~\ref{fig:backbone}) is implemented as a highly modular pre-norm Transformer~\cite{radford2019language, xiong2020layer} comprising $L$ configurable blocks. Rather than enforcing a static architecture, \textsc{DantinoX} exposes key structural choices as hyperparameter toggles. 

For the attention sublayer, the framework supports standard MHA~\cite{attention} as a reference baseline, alongside memory-efficient variants such as GQA~\cite{ainslie2023gqa} and MLA~\cite{liu2024deepseek}. All attention mechanisms (which can be \textit{causal} or \textit{bidirectional}, depending on the \textit{paradigm}) seamlessly integrate modern optimizations, including FlashAttention~\cite{dao2022flashattention}, sliding-window contexts~\cite{beltagy2020longformer}, gated~\cite{qiu2026gated}, linear~\cite{katharopoulos2020transformers} and differential attention~\cite{ye2025differential}. 
Similarly, the feed-forward network (FFN) can be instantiated as a standard dense MLP with SwiGLU~\cite{shazeer2020glu} or other activations like GELU~\cite{hendrycks2016gaussian}, a sparse Mixture-of-Experts (MoE)~\cite{shazeer2017outrageously, fedus2022switch} with top-$k$ routing, or the recent LatentMoE~\cite{elango2026latentmoe}. Users can select the positional encoding scheme (RoPE~\cite{su2024roformer}, sinusoidal~\cite{attention}, learned, or none) and the normalization layer (RMSNorm~\cite{zhang2019root} or LayerNorm~\cite{ba2016layer}). For downstream adaptation, LoRA adapters~\cite{hu2022lora} can be injected into any projection matrix. Additionally, weight tying~\cite{press2017using, inan2016tying} between input embeddings and output projections is natively supported for autoregressive and discrete diffusion modelling.

\begin{figure}[t] 
  \centering
\begin{tikzpicture}[
  node distance = 0.35cm,
  every node/.style = {font=\sffamily\scriptsize},
  box/.style = {
    draw=black, thick, fill=white, 
    minimum width=3.4cm, minimum height=0.45cm, align=center
  },
  note/.style = {
    draw=black!40, densely dotted, fill=gray!5, 
    align=left, font=\tiny\sffamily, inner sep=3pt
  },
  plus/.style = {
    circle, draw=black, thick, fill=white, 
    inner sep=0pt, minimum size=0.35cm, font=\small\bfseries
  },
  arr/.style = {-{Stealth[length=4pt, width=3pt]}, thick},
  skip/.style = {-{Stealth[length=4pt, width=3pt]}, thick, rounded corners=2pt}
]

\node[box] (in) {Token IDs $\mathbf{x} \in \mathbb{Z}^{B \times T}$};

\node[box, below=0.3cm of in] (emb) {Embedding + Pos. Enc.};
\node[note, right=0.25cm of emb] (embo) {RoPE $\mid$ Sinusoidal $\mid$ Learned};
\draw[densely dotted, gray, thick] (emb) -- (embo);

\node[box, below=0.5cm of emb] (norm1) {Pre-Norm (RMS $\mid$ LN)};

\node[box, below=0.3cm of norm1] (attn) {Self-Attention};
\node[note, right=0.25cm of attn] (attno) {
  MHA $\mid$ GQA $\mid$ MLA \\
  Flash $\cdot$ Sliding $\cdot$ Diff \\   Gated $\cdot$ Linear
};
\draw[densely dotted, gray, thick] (attn) -- (attno);

\node[plus, below=0.3cm of attn] (add1) {$+$};

\node[box, below=0.4cm of add1] (norm2) {Pre-Norm (RMS $\mid$ LN)};

\node[box, below=0.3cm of norm2] (ffn) {Feed-Forward};
\node[note, right=0.25cm of ffn] (ffno) {
  Dense (MLP, SwiGLU) \\
  Sparse (MoE Top-$k$) \\
  Sparse (MoE Latent Top-$k$)
};
\draw[densely dotted, gray, thick] (ffn) -- (ffno);

\node[plus, below=0.3cm of ffn] (add2) {$+$};

\node[box, below=0.4cm of add2] (out) {$\mathbf{h}\!\in\!\mathbb{R}^{B\times T\times D}$\enspace
   $\to$\enspace backbone head};

\draw[arr] (in) -- (emb);
\draw[arr] (emb) -- (norm1);
\draw[arr] (norm1) -- (attn);
\draw[arr] (attn) -- (add1);
\draw[arr] (add1) -- (norm2);
\draw[arr] (norm2) -- (ffn);
\draw[arr] (ffn) -- (add2);
\draw[arr] (add2) -- (out);

\coordinate (s1) at ($(emb.south)!0.35!(norm1.north)$);
\coordinate (s2) at ($(add1.south)!0.4!(norm2.north)$);

\fill (s1) circle (1.2pt);
\fill (s2) circle (1.2pt);

\draw[skip] (s1) -- ++(-2.2,0) |- (add1);
\draw[skip] (s2) -- ++(-2.2,0) |- (add2);

\begin{pgfonlayer}{background}
  \coordinate (f_top) at ($(in.south)!0.4!(emb.north)$);
  \coordinate (f_bot) at ($(add2.south) + (0, -0.10)$);
  
  \coordinate (f_left) at (-2.6, 0); 
  \coordinate (f_right) at ($(attno.east) + (0.15, 0)$);
  
  \node[
    draw=black!60, thick, dashed, fill=gray!3, inner sep=0pt,
    fit=(f_left |- f_top) (f_right |- f_bot)
  ] (frame) {};
\end{pgfonlayer}
\node[font=\scriptsize\bfseries, text=black, fill=white, inner sep=2pt] 
  at ($(frame.north west)!0.3!(frame.north)$) {};

\node[anchor=south east, font=\small\bfseries, text=black, inner sep=4pt] 
  at (frame.south east) {$\times L$};
  
\end{tikzpicture}
\caption{Unified Backbone.}
    \label{fig:backbone}
\end{figure}
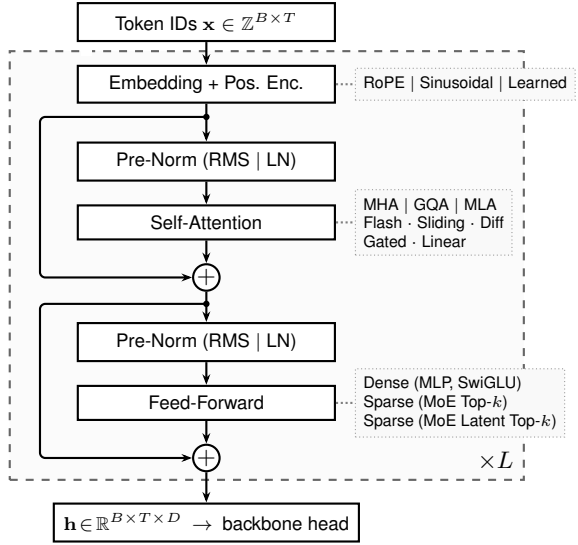

\begin{figure*}[t] 
    \centering
    \begin{tikzpicture}[
      node distance = 0.4cm and 0.5cm,
      every node/.style = {font=\scriptsize},
      mb/.style = {draw=#1!80!black, fill=#1!10, rounded corners=3pt,
                   minimum width=3.8cm, minimum height=0.6cm,
                   align=center, inner sep=2.5pt},
      mb/.default = corecol,
      arr/.style = {-{Stealth[length=5pt,width=3.5pt]}, thick, color=#1},
      arr/.default = corecol,
    ]

    \node[mb=corecol, minimum width=8cm] (Root) 
      {\textbf{$\mathbf{h}\!\in\!\mathbb{R}^{B\times T\times D}$\enspace
        $\to$\enspace Backbone Head}};

    \node[mb=disccol, below=0.60cm of Root, 
          label={[font=\tiny, text=optcol, xshift=35pt, yshift=-1pt]north west:Training}] (Adisc)
      {\textbf{Discrete}\;(bidirectional)\\\scriptsize masked cross-entropy};
      
    \node[mb=arcol, left=0.4cm of Adisc,
          label={[font=\tiny, text=optcol, xshift=35pt, yshift=-1pt]north west:Training}] (Aar)
      {\textbf{Autoregressive}\;(causal)\\\scriptsize next-token cross-entropy};
      
    \node[mb=elfcol, right=0.4cm of Adisc,
          label={[font=\tiny, text=optcol, xshift=35pt, yshift=-1pt]north west:Training}] (Aelf)
      {\textbf{Continuous}\;(bidirectional)\\\scriptsize MSE\,+\,CE in $\mathbb{R}^E$};

    \node[mb=arcol, below=0.35cm of Aar,
          label={[font=\tiny, text=optcol, xshift=35pt, yshift=-1pt]north west:Inference}] (Aar2)
      {KV-Cache / No Cache decode\\\tiny greedy\,/\,top-$k$\,/\,top-$p$};

    \node[mb=disccol, below=0.35cm of Adisc,
          label={[font=\tiny, text=optcol, xshift=35pt, yshift=-1pt]north west:Inference}] (Adisc2)
      {Reverse diffusion\;($S$\;steps)\\\tiny (sample\,/\,confidence\,/\,factor) | blocks};

    \node[mb=elfcol, below=0.35cm of Aelf,
          label={[font=\tiny, text=optcol, xshift=35pt, yshift=-1pt]north west:Inference}] (Aelf2)
      {ODE\,/\,SDE\;($n$\;steps)\\\tiny CFG scale $w$,\;noise $\gamma$};

    \draw[arr=arcol]   (Root.south) -- ++(0,-.25) -| (Aar.north);
    \draw[arr=disccol] (Root.south) -- (Adisc.north);
    \draw[arr=elfcol]  (Root.south) -- ++(0,-.25) -| (Aelf.north);

    \draw[arr=arcol]   (Aar)   -- (Aar2);
    \draw[arr=disccol] (Adisc) -- (Adisc2);
    \draw[arr=elfcol]  (Aelf)  -- (Aelf2);

    \end{tikzpicture}
    \caption{Generative paradigms and corresponding inference strategies.}
    \label{fig:paradigms}
\end{figure*}
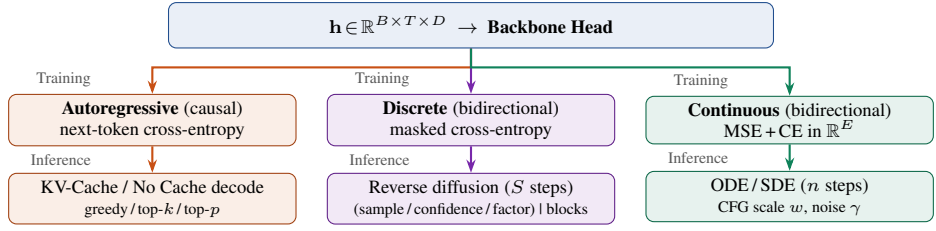

\subsection{Generation Paradigms}
Figure~\ref{fig:paradigms} illustrates how each paradigm head mounts on the shared backbone
output $\mathbf{h} \in \mathbb{R}^{B \times T \times D}$ and how training
and inference differ across the three modes.

\noindent\textbf{Autoregressive (AR).} 
Under the AR paradigm, the backbone employs a standard causal attention mask~\cite{attention} and is optimized by minimizing the next-token cross-entropy. During inference, generation is accelerated via KV-cache~\cite{shazeer2019fast, pope2023efficiently} to prevent context recomputation. The generator supports different decoding strategies, including greedy search, temperature scaling, top-$k$~\cite{fan2018hierarchical}, and top-$p$ sampling~\cite{holtzman2019curious}.


\noindent\textbf{Discrete Diffusion.}
For discrete diffusion, \textsc{DantinoX} implements the LLaDA formulation~\cite{nie2024llada}, building upon foundational masked diffusion language models~\cite{austin2021structured, sahoo2024mdlm}. In this mode, the backbone utilizes \textit{bidirectional} attention. During training, tokens are independently replaced by a \texttt{[MASK]} token according to a time $t$ configurable schedule (linear, cosine, or square-root)~\cite{austin2021structured} and optimized via a $(1/t)$-weighted cross-entropy loss over the masked positions. Inference is performed through $S$ reverse-diffusion steps~\cite{ho2020denoising}, with built-in support for multiple decoding strategies (\texttt{sample}, \texttt{greedy}, \texttt{confidence}, and \texttt{factor}). The framework also provides a \textit{block-generation} mode~\cite{nie2024llada, wu2025fast}, which denoises local windows sequentially and, to optimize throughput, a \textit{DualCache}~\cite{wu2025fast} for a greater speedup with low accuracy loss.

\noindent\textbf{Continuous Flow-Matching.}
For continuous flow-matching, \textsc{DantinoX} adopts the recent ELF formulation~\cite{hu2026elf}. This approach operates within the continuous embedding space of a frozen encoder (e.g., T5~\cite{raffel2020exploring}), avoiding per-step token-level supervision by discretizing (decoding) only at the final denoising step via a shared-weight network. During training, the backbone learns the \textit{time-dependent vector field}~\cite{lipman2022flow} corresponding to the forward process $\mathbf{z}_t = t\,\mathbf{x} + (1{-}t)\,\boldsymbol\varepsilon$ (where $\mathbf{x}$ is the normalized embedding, $\boldsymbol\varepsilon \sim \mathcal{N}(\mathbf{0},\mathbf{I})$, and $t \sim \mathcal{U}[0,1]$), and is optimized via a combined MSE and cross-entropy loss. Inference relies on standard ODE or SDE integration, with native support for Classifier-Free Guidance~\cite{ho2022classifier} modulated by a \texttt{cfg\_scale} parameter.

\begin{figure}[t]
\begin{lstlisting}[style=dxpy]
import dantinox as dx

# 1. BackBone configuration 
cfg = dx.ModelConfig(
    paradigm   = "discrete",
    attention  = "gqa", kv_heads = 4, # "mha"|"gqa"|"mla"
    ffn        = "mlp", use_swiglu = True, # mlp | moe
    dim        = 512,   n_heads    = 16,
    num_blocks = 12, 
)

# 2. Training configuration
tcfg = dx.TrainingConfig(
    lr = 3e-4, epochs = 10, batch_size = 64, grad_accum = 4,  
    optimizer = "muon", lr_schedule = "cosine", warmup_steps = 400, n_devices = 4, tp_size = 2, tokenizer_type="bpe", gradient_checkpoint="True"
)

# 3. Training
run_dir = dx.Trainer(dx.Paradigm(cfg), tcfg).fit(dataset)

# 4. Streaming Inference (auto-dispatches to paradigm)
gen = dx.Generator(run_dir, seed=42)
for chunk in gen.stream(prompt, max_new_tokens=100, n_steps=50, decoding_strategy="confidence"):
    print(chunk, end="", flush=True)


\end{lstlisting}
\vspace{-2mm} 
\caption{\textsc{DantinoX} API.
\textbf{(1)}~Training a discrete diffusion model via a single
\texttt{ModelConfig}
\textbf{(2)}~Paradigm-agnostic streaming inference via
\texttt{Generator.stream}.}
\label{lst:api_usage}
\end{figure}

\subsection{Training and Testing}

\textbf{Training.} \textsc{DantinoX} provides a unified training pipeline that entirely abstracts the specificities of the chosen generation paradigm. Users can seamlessly couple a configured backbone with their target paradigm and execute the entire training process via a single \texttt{fit()} call (Figure~\ref{lst:api_usage}). Within the \texttt{TrainingConfig}, users can dynamically specify the desired tokenization scheme, such as character-level, SentencePiece~\cite{kudo-richardson-2018-sentencepiece}, or Byte-Pair Encoding (BPE)~\cite{sennrich-etal-2016-neural}, and the underlying model automatically adjusts its vocabulary and embedding matrix accordingly. The training loop integrates the \texttt{Optax} ecosystem\footnote{\url{https://optax.readthedocs.io/}}, supporting standard optimization algorithms as well as recent advancements like Muon\footnote{\url{https://kellerjordan.github.io/posts/muon/}}. This is complemented by standard stability and efficiency utilities, including cosine~\cite{loshchilov2016sgdr} and linear learning-rate schedules with warm-up~\cite{goyal2017accurate}, gradient accumulation, gradient clipping~\cite{pascanu2013difficulty}, and \texttt{bfloat16} mixed precision~\cite{kalamkar2019study}. For distributed execution, the framework natively handles multi-GPU scaling. By simply defining the desired device topology, users can freely compose Data Parallelism (DP) and JAX SPMD-based Tensor Parallelism (TP)~\cite{adler2024nemotron} to efficiently distribute the workload.

\textbf{Inference and Benchmarking.}
As illustrated in Figure~\ref{lst:api_usage}, the \texttt{Generator} module parses the checkpoint configuration and automatically routes execution to the appropriate paradigm. 
Beyond generation, \textsc{DantinoX} provides dedicated tools for rigorous model profiling. The \texttt{dx.count\_flops} utility enables zero-execution profiling by analytically computing the FLOP breakdown per architectural component, while \texttt{dx.profile} captures runtime latency and Model FLOP Utilization (MFU). For system-level evaluation, \texttt{BenchmarkSuite.default()} automates comprehensive grid sweeps across varying batch sizes and sequence lengths. In a single run, it evaluates throughput, latency, and perplexity utilized for the analyses in Section~4.

\section{Evaluation}

We validate \textsc{DantinoX} on three fronts, all driven by the same
public pipeline with no paradigm-specific code. We first check the correctness of the
paradigm implementations against independent codebases~\cite{zhou2026dllm, xlm2026eacl}
(Section~\ref{sec:external_validation}). We then compare generation
quality across all nine paradigm~$\times$~attention combinations. To ensure a rigorous evaluation, we maintain identical data, tokenizers, optimizers, and computational budgets (Section~\ref{sec:gen_quality}). Finally, using the \textsc{DantinoX} benchmark suite, we profile inference latency, throughput, and energy across paradigms through a roofline analysis (Section~\ref{sec:eval:inference}).

\begin{figure}[t]
\centering
\begin{subfigure}{0.48\textwidth}
\includegraphics[width=\textwidth]{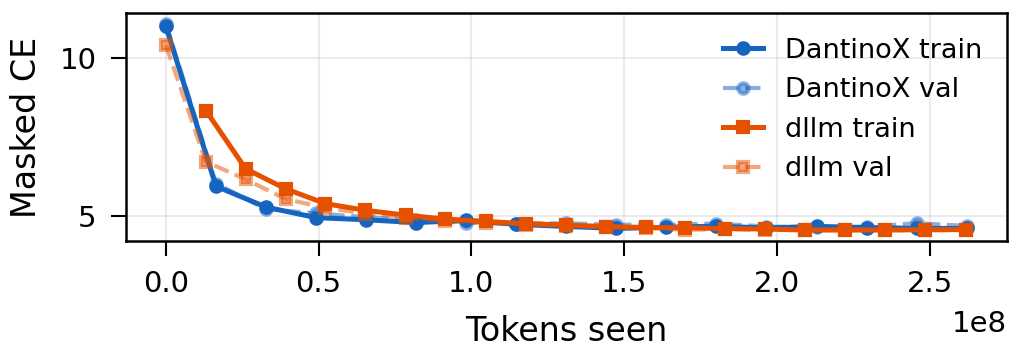}
\caption{\textsc{DantinoX} Diffusion \& \texttt{dllm}}
\end{subfigure}
\begin{subfigure}{0.48\textwidth}
\includegraphics[width=\textwidth]{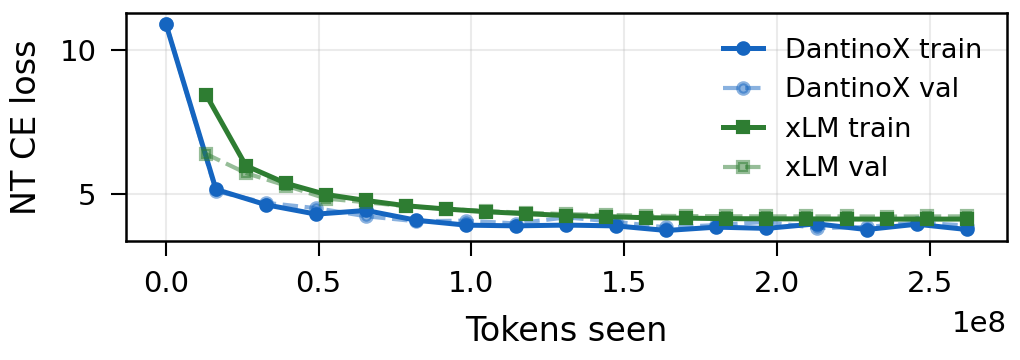}
\caption{\textsc{DantinoX} AR \& \texttt{xLM}}
\end{subfigure}
\caption{Train/val loss vs.\ tokens seen, \textsc{DantinoX} vs.\ \texttt{dllm} and \texttt{xLM}
  PyTorch implementations.}
\label{fig:cross_framework_validation}
\end{figure}

\subsection{External Validation}
\label{sec:external_validation}

To verify that DantinoX's paradigm implementations are not artefacts of a single codebase, we reproduce its discrete-diffusion and autoregressive training with independent
implementations, \texttt{dllm}~\cite{zhou2026dllm} (LLaDA architecture) and \texttt{xLM}~\cite{xlm2026eacl} (AR architecture).
Both comparisons match, on both sides: architecture (512-d, 8~attention heads, 12~layers, plain GELU FFN, T5 SentencePiece vocabulary of
32{,}128); loss formula (the $(1/t)$-weighted masked cross-entropy for diffusion, standard next-token cross-entropy for AR); optimizer (\texttt{AdamW}, identical learning rate, warmup, and cosine schedule); and data (WikiText-103-raw-v1, 50M-token cap, 90/10 train/validation split, 262M-token budget).
As Figure~\ref{fig:cross_framework_validation} shows, the trajectories track each other closely, with final losses within $1\%$ for diffusion and $8\%$ for AR (the residue consistent with defaults \textsc{DantinoX} retains, such as RMSNorm and weight tying, vs.\ xLM's LayerNorm and
untied embeddings). Matching xLM's FFN required only flipping
\textsc{DantinoX}'s \texttt{use\_swiglu} flag; restoring the
SwiGLU recovers a further $0.14$~nats of validation loss, an ablation
that in \texttt{xLM} codebase would require code modifications.

\subsection{Open-Ended Generation Quality}
\label{sec:gen_quality}

\begin{table}[t]
\centering
\renewcommand{\arraystretch}{1.0} 
\setlength{\tabcolsep}{3pt}
\resizebox{\columnwidth}{!}{%
\begin{tabular}{@{} l rrrrr @{}}
\toprule
\textbf{Arch (Params)}
  & MV~$\uparrow$ & PPL~$\downarrow$
  & D-2~$\uparrow$ & R-4~$\downarrow$ & B-4~$\uparrow$ \\
\midrule
%
\rowcolor{gray!10}
\multicolumn{6}{@{}l}{\textbf{Autoregressive}} \\
\quad MHA (67M)  & 0.17{\tiny$\pm$.03} & \textbf{1216}{\tiny$\pm$5} & \textbf{0.697}{\tiny$\pm$.000} & 0.007{\tiny$\pm$.000} & 0.052{\tiny$\pm$.000} \\
\quad GQA (62M)  & 0.18{\tiny$\pm$.06} & 1233{\tiny$\pm$3}          & 0.688{\tiny$\pm$.001}          & 0.005{\tiny$\pm$.000} & 0.050{\tiny$\pm$.004} \\
\quad MLA (65M)  & 0.07{\tiny$\pm$.01} & 1860{\tiny$\pm$3}          & 0.682{\tiny$\pm$.001}          & \textbf{0.003}{\tiny$\pm$.000} & 0.020{\tiny$\pm$.002} \\
\midrule
%
\rowcolor{gray!10}
\multicolumn{6}{@{}l}{\textbf{Discrete Diffusion}} \\
\quad MHA (70M)  & 0.20{\tiny$\pm$.04} & 1834{\tiny$\pm$43} & 0.728{\tiny$\pm$.006} & 0.019{\tiny$\pm$.010} & 0.029{\tiny$\pm$.000} \\
\quad GQA (65M)  & 0.12{\tiny$\pm$.00} & 1777{\tiny$\pm$57} & 0.733{\tiny$\pm$.003} & \textbf{0.006}{\tiny$\pm$.000} & 0.030{\tiny$\pm$.003} \\
\quad MLA (70M)  & 0.15{\tiny$\pm$.05} & 1803{\tiny$\pm$1}  & 0.720{\tiny$\pm$.008} & 0.020{\tiny$\pm$.009} & 0.033{\tiny$\pm$.003} \\
\midrule
%
\rowcolor{gray!10}
\multicolumn{6}{@{}l}{\textbf{Continuous Flow-Matching}} \\
\quad MHA (82M)  & 0.80{\tiny$\pm$.04} & 234.6{\tiny$\pm$2.0}          & \textbf{0.627}{\tiny$\pm$.001} & \textbf{0.007}{\tiny$\pm$.000} & --- \\
\quad GQA (77M)  & 0.69{\tiny$\pm$.09} & 188.0{\tiny$\pm$0.4}          & 0.562{\tiny$\pm$.001}          & 0.027{\tiny$\pm$.000}          & --- \\
\quad MLA (79M)  & 0.78{\tiny$\pm$.07} & \textbf{156.3}{\tiny$\pm$0.3} & 0.538{\tiny$\pm$.001}          & 0.107{\tiny$\pm$.000}          & --- \\
\bottomrule
\end{tabular}%
}
\vspace{-2mm} 
\caption{%
  \textsc{DantinoX} open-ended generation quality at Small scale
  (512-d, 12-layer).
}
\label{tab:genquality}
\end{table}

To demonstrate the end-to-end evaluation capabilities of \textsc{DantinoX}, we present a controlled, small-scale cross-paradigm comparison: all nine paradigm~$\times$~attention combinations are trained and evaluated through the same pipeline, varying only the configuration paradigm field, as shown in Fig.~\ref{lst:table2}. 
All models share the training recipe (WikiText-103~\cite{merity2016pointer},
\texttt{Muon} optimizer, effective batch of 256~sequences of 512~tokens) and the
same \texttt{SentencePiece} tokenizer~\cite{kudo-richardson-2018-sentencepiece}
(T5~vocabulary), so that differences in outcomes can be attributed to
the paradigm rather than to implementation details.
We measure four complementary axes of generation quality
(Table~\ref{tab:genquality}): MAUVE~\cite{pillutla2021mauve}
(human-likeness), PPL\textsubscript{GPT-2} (fluency),
Distinct-2 and Rep-4~\cite{li2016diversity,welleck2019neural}
(diversity and repetition), and
BLEU-4\textsubscript{cond}~\cite{papineni-etal-2002-bleu}
(conditional continuation), jointly, because fluency metrics alone
reward repetitive text.
Results average 100~unconditional samples of 128~tokens over three
generation seeds, with a matched inference budget of 64~steps per
paradigm. BLEU-4\textsubscript{cond} is not reported for Flow-Matching: while the
ELF formulation supports prefix conditioning~\cite{hu2026elf}, our
current implementation does not yet expose it; closing this gap is
planned future work.
At this scale (${\sim}$65--80M parameters, single-domain data), the
comparison reveals a clear division of labour: Flow-Matching produces
the most fluent text ( margin reflects that its frozen-encoder embedding space yields more coherent short samples at this scale, not a difference in measurement), Diffusion the most lexically diverse, and AR the
most accurate conditional continuations with the least repetition, 
each gap an order of magnitude above seed variance.
We also observe that the effect of attention is paradigm-dependent:
MLA is the most fluent variant in Flow-Matching (at a heavy repetition
cost) yet the weakest in AR. These absolute numbers reflect a deliberately small compute budget; the contribution is not the ranking but that \textsc{DantinoX} makes such a controlled configuration changes alone.
We make no claim that these specific rankings persist at larger scales
or on broader corpora,  establishing that is precisely the kind of
question \textsc{DantinoX} is designed to make cheap to answer.
What the study validates is the workflow: nine controlled
paradigm~$\times$~attention comparisons, with seed-level variance
reporting, produced entirely by varying configuration fields.

\begin{figure}[t!]
    \centering
    \begin{subfigure}{\columnwidth}
        \includegraphics[width=\linewidth]{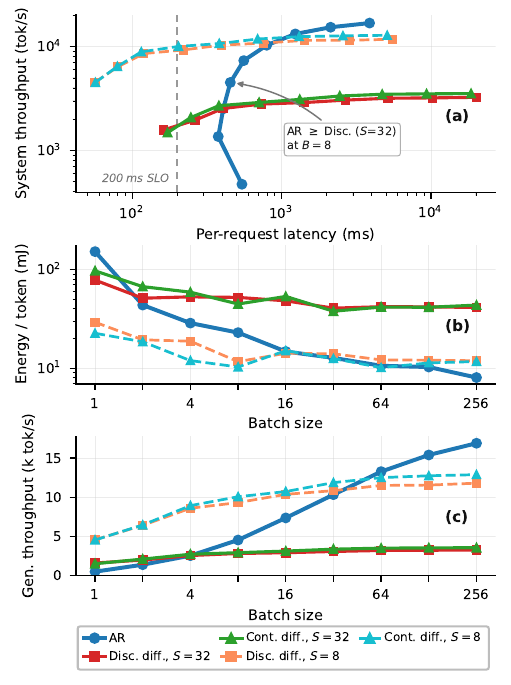}
        \caption{Inference efficiency of AR, discrete diffusion, and continuous flow-matching.}
        \label{fig:inference_efficiency}
    \end{subfigure}
    
    \vspace{2mm} 
    
    \begin{subfigure}{\columnwidth}
        \includegraphics[width=\linewidth]{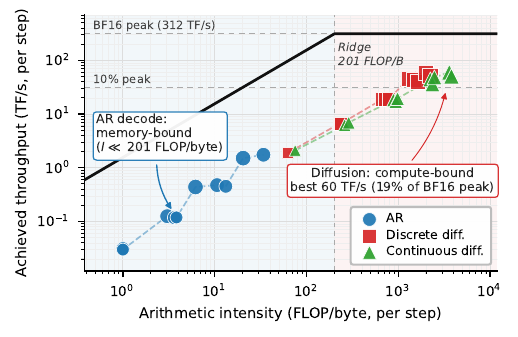}
        \caption{Hardware utilization roofline (A100 GPU, \texttt{bf16}).}
        \label{fig:roofline}
    \end{subfigure}
    
    \caption{Inference performance on the Large backbone (1024-d, 16-layer, $\approx$130\,M parameters).}
    \label{fig:inference_evaluation}
\end{figure}

\subsection{Inference Efficiency Across Paradigms}
\label{sec:eval:inference}

This study demonstrates \textsc{DantinoX}'s integrated benchmarking suite: a
single call to \texttt{BenchmarkSuite.default()} sweeps latency,
throughput, and energy for all three paradigms on one shared backbone
(1024-d, 16-layer, ${\approx}$130M parameters; A100-40GB, \texttt{bf16},
$G=256$ generated tokens, $B \in [1,256]$, diffusion at
$S \in \{8,32\}$ steps), and the framework's profiling tools
(\texttt{count\_flops}, \texttt{profile}) ground the results in a
roofline analysis (Figure~\ref{fig:roofline}).
The sweep reveals that no paradigm is most efficient everywhere;
instead, AR and diffusion trade places at a well-defined crossover
(Figure~\ref{fig:inference_efficiency}).
At low batch sizes, diffusion dominates interactive serving
(Figure~\ref{fig:inference_efficiency}): by refining
all positions jointly over $S$ parallel passes, it completes a single
request in 56\,ms where token-by-token AR decoding takes 542\,ms,
past the 200\,ms interactive SLO~\cite{miller1968response}, at
roughly $7\times$ lower energy per token.
As batch size grows, the ranking reverses: KV-cache amortisation raises
AR's arithmetic intensity across concurrent sequences, and AR overtakes
all diffusion variants by $B \approx 32$, with the highest throughput
and lowest energy at $B=256$.
The roofline analysis (Figure~\ref{fig:roofline}) explains the
asymmetry: AR decoding is memory-bound and thus gains the most from
batching, whereas each diffusion step is already compute-bound on this
hardware, so extra concurrency yields diminishing returns.
For practitioners, this yields a concrete decision rule: deploy diffusion
at low $S$, subject to the trade-off between denoising steps and sample quality, when serving few
concurrent users under strict latency constraints, and
switch to AR for sustained, high-throughput serving at larger scales.
Because all paradigms share the same \texttt{ModelConfig} API within
\textsc{DantinoX}, shifting between these two deployment regimes requires
zero inference code changes.

\section{Conclusion}

In this paper, we introduced \textsc{DantinoX}, an open-source JAX/Flax framework that unifies language modeling across three major paradigms: autoregressive decoding, discrete masked diffusion, and continuous flow-matching. Previously, evaluating these methods required navigating completely separate codebases, making fair comparisons difficult. By integrating them into a single modular Transformer backbone, \textsc{DantinoX} solves this problem. Users can now switch between generative paradigms, attention mechanisms (such as MHA, GQA, and MLA), and hardware setups simply by changing a configuration file.
By making multi-paradigm training and streaming inference highly accessible through a single API, \textsc{DantinoX} is designed to support a diverse community. The framework simplifies complex ablations without requiring code edits, provides a cohesive environment for teaching modern NLP, and offers reusable pipelines that eliminate the need to build custom infrastructure. We hope \textsc{DantinoX} serves as a reliable, zero-boilerplate foundation for researchers, educators, and practitioners to build and evaluate the next generation of language models.

\section*{Broader Impact Statement}

We offer \textsc{DantinoX} with the hope of positively impacting the natural language processing community by making multi-paradigm generative modeling more accessible.

While discrete diffusion and continuous flow-matching are rapidly emerging as viable alternatives to autoregressive decoding, comparing these paradigms remains difficult due to fragmented codebases.
By providing a unified, open-source Transformer backbone, our framework aims to facilitate fair and controlled comparisons, helping researchers isolate true algorithmic differences from mere implementation details.
Beyond academic research, we hope this shared infrastructure proves useful to educators teaching modern generative techniques and to practitioners seeking reliable pipelines without having to build them from scratch.

Additionally, we recognize the environmental and computational costs associated with training language models. To support more mindful development, we integrated hardware profiling and energy-tracking tools directly into the framework, aiming to encourage resource-aware research.
Finally, because any generative infrastructure can be misused to produce biased or harmful text, we encourage developers who build upon our work to carefully evaluate their models for safety and ethical alignment before practical deployment.



\bibliography{custom}

\clearpage
\appendix

\section{Appendix}
\label{sec:appendix}

\textbf{Generation Quality and Inference Profiling Pipeline.}
Figure~\ref{lst:table2} details the automated pipeline used to generate the open-ended generation quality results presented in Table~\ref{tab:genquality}. It demonstrates how \textsc{DantinoX} enables a seamless sweep across the $3 \times 3$ grid of generation paradigms and attention mechanisms. By entirely decoupling the model configuration and the training loop from the specific generative formulation, researchers can conduct strictly controlled comparisons using generic text-scoring utilities without any paradigm-specific boilerplate.
Figure~\ref{lst:pipeline} provides the measurement script utilized for the inference efficiency and hardware roofline analyses discussed in Section~\ref{sec:eval:inference}. The snippet illustrates the use of \textsc{DantinoX}'s built-in profiling tools, such as \texttt{LatencyMetric}, \texttt{EnergyMetric}, and \texttt{FLOPsMetric}, to systematically benchmark autoregressive decoding, discrete diffusion, and continuous flow-matching. This demonstrates how different scaling behaviours across varying batch sizes and generation constraints can be profiled through a single, unified API.

\begin{figure}[h!]
\begin{lstlisting}[style=dxpy]
import dantinox as dx

GRID    = [(p, a) for p in ("ar", "discrete", "continuous")
                   for a in ("mha", "gqa", "mla")]
N_STEPS = 64   # matched inference budget across paradigms

rows = []
for paradigm, attn in GRID:
    for seed in (42, 43, 44):
        cfg  = dx.ModelConfig(paradigm=paradigm, attention=attn, dim=512, n_heads=8, num_blocks=12)
        
        tcfg = dx.TrainingConfig(optimizer="muon",             batch_size=256,  dataset_name="wikitext", seed=seed)
        
        run_dir = dx.Trainer(dx.Paradigm(cfg), tcfg).fit(
            dataset_source="huggingface")

        gen = dx.Generator(run_dir, seed=seed)
        
        samples = [gen.generate("", max_new_tokens=128, n_steps=N_STEPS) for _ in range(100)] # gen.seed varied per call
        
        rows.append(score(samples, paradigm))   # MAUVE, PPL, D-2, R-4, B-4

\end{lstlisting}
\vspace{-3mm}
\caption{\textsc{DantinoX} pipeline reproducing Table~\ref{tab:genquality}.}
\label{lst:table2}
\end{figure}

\begin{figure}[h!]
\begin{lstlisting}[style=dxpy]
import dantinox as dx, jax.numpy as jnp
from dantinox.profiling import count_flops, LatencyMetric, FLOPsMetric, EnergyMetric
from flax import nnx

ARCH = dict(dim=1024, n_heads=16, num_blocks=16, vocab_size=32128)
SERIES = [("ar", {"use_cache": True}), ("discrete", {"n_steps": 32}), 
          ("discrete", {"n_steps": 8}), ("continuous", {"n_steps": 32}), 
          ("continuous", {"n_steps": 8})]

for par, kw in SERIES:
    cfg = dx.ModelConfig(paradigm=par, **ARCH)
    p = dx.Paradigm(cfg)
    m = p.build_model(rngs=nnx.Rngs(0))
    for B in [1, 4, 16, 64, 128, 256]:
        T = B * 128
        fn = lambda: p.generate(m, jnp.ones((B, 128), jnp.int32), rng, **kw)
        lat = LatencyMetric(n_warmup=5, n_measure=50).measure(fn, n_tokens=T)
        eng = EnergyMetric().measure(fn, n_tokens=T)
        bd = count_flops(cfg, seq_len=128, batch_size=B)
        mfu = FLOPsMetric(312.0).measure(cfg, 128, B, elapsed_s=lat.mean_ms/1e3)
\end{lstlisting}
\vspace{-3mm}
\caption{\textsc{DantinoX} pipeline for Figs.~\ref{fig:inference_efficiency} and~\ref{fig:roofline}.}
\label{lst:pipeline}
\end{figure}

\begin{figure}[htbp]
    \centering
    
    \begin{subfigure}[b]{\columnwidth}
        \centering
        \includegraphics[width=\textwidth]{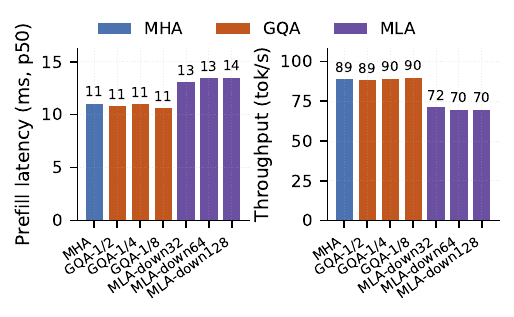}
        \caption{\footnotesize \textbf{AR}: Prefill latency and decode throughput per attention variant.}
        \label{fig:appx_attention_type}
    \end{subfigure}
    
    \vspace{4mm} 
    
    \begin{subfigure}[b]{\columnwidth}
        \centering
        \includegraphics[width=\textwidth]{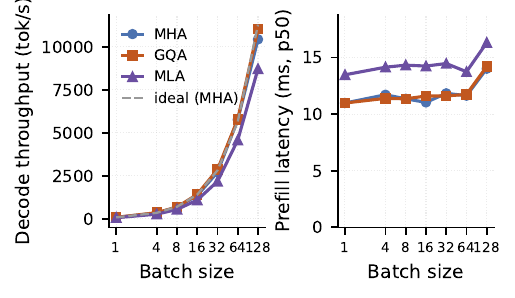}
        \caption{\footnotesize \textbf{AR}: Decode throughput and prefill latency vs.\ batch size.}
        \label{fig:appx_batch}
    \end{subfigure}
    
    \vspace{4mm}
    
    \begin{subfigure}[b]{\columnwidth}
        \centering
        \includegraphics[width=\textwidth]{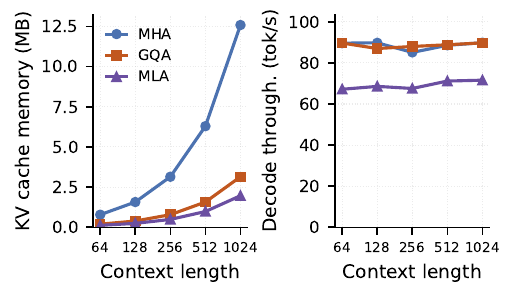}
        \caption{\footnotesize \textbf{AR}: KV-cache memory and decode throughput vs.\ context length.}
        \label{fig:appx_context}
    \end{subfigure}
    \caption{\footnotesize Extended inference benchmarks isolating the impact of individual attention variants (MHA, GQA, MLA) on autoregressive generation performance and memory footprint.}
\end{figure}

\begin{figure}[t]
\centering
\includegraphics[width=\columnwidth]{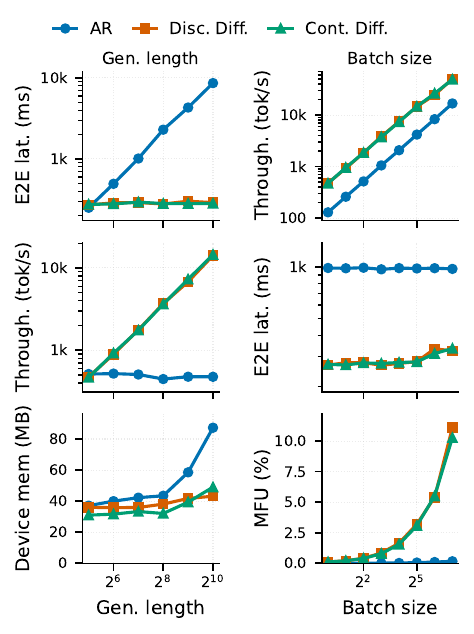}
\caption{\footnotesize Latency, throughput, and memory/MFU vs.\
  generation length (left) and batch size (right), per paradigm.}
\label{fig:appx_scaling}
\end{figure}

\textbf{Extended Inference Benchmarks.}
We complement Section~\ref{sec:eval:inference} with finer-grained
sweeps: \textit{autoregressive} (AR) attention-variant characterization in isolation
(Figures~\ref{fig:appx_attention_type}, \ref{fig:appx_context}, using \texttt{char} tokenizer), and paradigm scaling with generation length and batch size
(Figure~\ref{fig:appx_scaling}, shown for MHA; GQA and MLA are
qualitatively identical).
Figure~\ref{fig:appx_attention_type} isolates the attention variant on
an 8M backbone (B{=}1, fp32): GQA matches MHA at every compression
ratio, while MLA trades ${\approx}20\%$ decode throughput for its
compressed cache. Figure~\ref{fig:appx_batch} scales batch size: MHA
and GQA track ideal linear throughput scaling to B{=}128, MLA scales
linearly at a ${\approx}20\%$ lower slope, and prefill latency stays
batch-insensitive until B{=}128. Figure~\ref{fig:appx_context} scales
context length: MHA's KV cache grows ${\approx}4\times$ faster than
GQA-1/4's and ${\approx}6\times$ faster than MLA's, the memory
saving that motivates both variants, while decode throughput
remains context-insensitive in this range.
Figure~\ref{fig:appx_scaling} (left) scales generation length (medium
backbone, B{=}4, $S{=}32$ denoising steps): AR latency grows linearly
with generated length, while both diffusion paradigms finish in a
fixed $S$ passes, so their latency stays flat and throughput grows
linearly instead. The same figure (right) scales batch size on the
same paradigms (medium ${\approx}6.6$M backbone, G{=}128, fp32): at
this scale every decode step is dispatch- rather than compute-bound,
so diffusion's $S{=}32$ parallel passes beat AR's 128 sequential
steps at every batch size shown, and the AR crossover of
Figure~\ref{fig:inference_efficiency} sits beyond this range. MFU
confirms the regime: ${\approx}11\%$ for diffusion vs.\ ${<}1\%$ for
AR.


\end{document}